\documentclass[runningheads]{llncs}
\usepackage{graphicx}
\usepackage{booktabs}
\usepackage{xcolor}
\usepackage{multirow}
\usepackage{caption}
\usepackage{amssymb}
\usepackage{amsmath}
\definecolor{paperYellow}{RGB}{255, 255, 230} 
\definecolor{paperGreen}{RGB}{230, 255, 230}  
\definecolor{paperRed}{RGB}{255, 230, 230}    

\begin{document}
\title{You Only Gaussian Once: Controllable 3D Gaussian Splatting for Ultra-Densely Sampled Scenes}
\titlerunning{YOGO}
%
\author{Jinrang Jia \and
Zhenjia Li \and
Yifeng Shi\thanks{Corresponding author}}
\authorrunning{J. Jin et al.}

\institute{KE Holdings Inc., Beijing, China\\
\email{shiyifeng@tju.edu.cn}}
\maketitle              
%

\begin{center}
  \includegraphics[width=\linewidth]{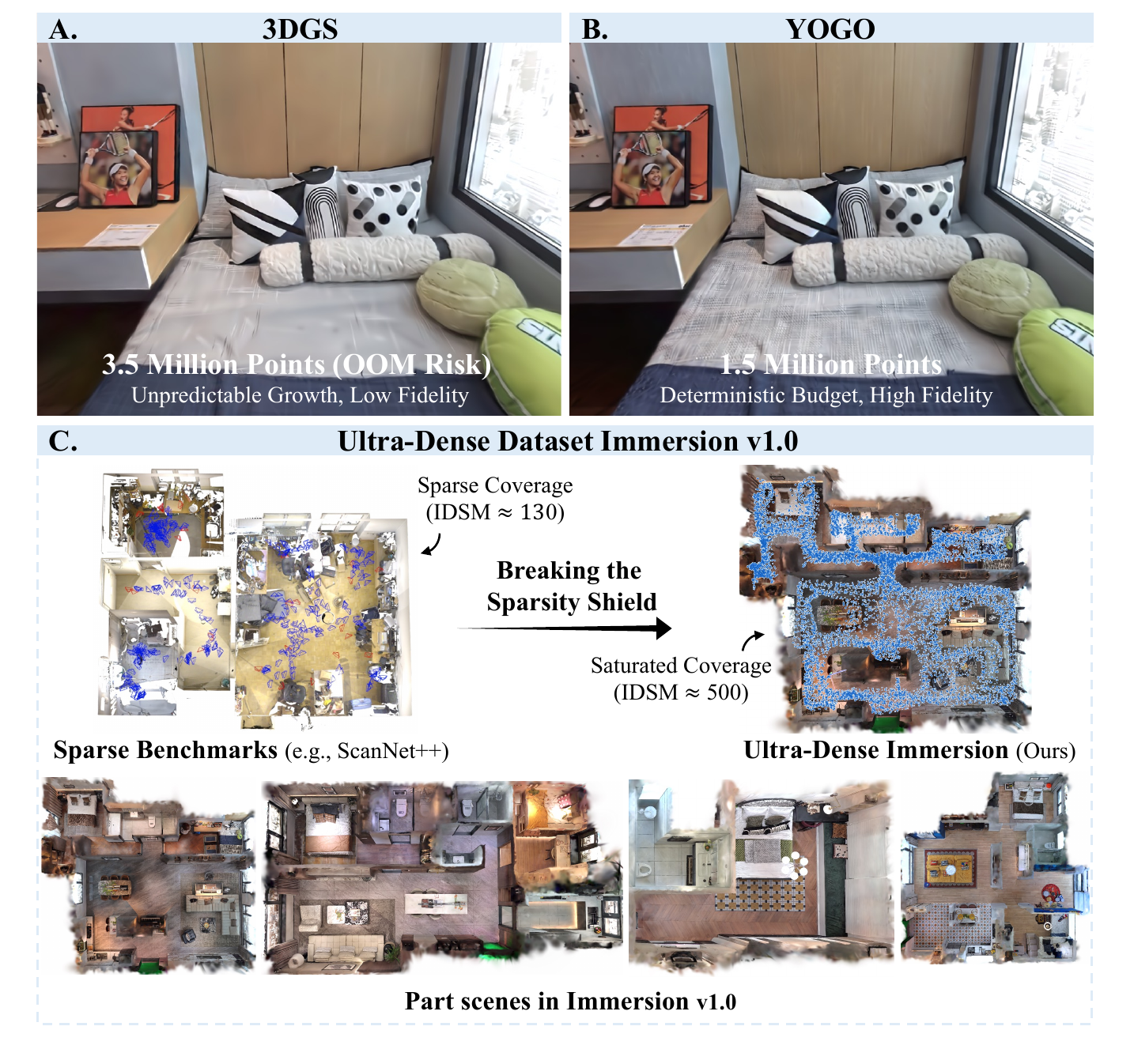} 
  \captionof{figure}{(A) \textbf{Vanilla 3DGS} suffers from uncontrollable growth and OOM risks on our challenging Immersion dataset. (B) \textbf{YOGO} ensures high-fidelity reconstruction under a deterministic budget (e.g., 1.5M points) via robust multi-sensor fusion. (C) Unlike sparse conventional benchmarks (left), Immersion provides ultra-dense saturation (right), breaking the sparsity shield to force true physical fidelity.}
  \label{fig:teaser}
\end{center}

\begin{abstract}
3D Gaussian Splatting (3DGS) has revolutionized neural rendering, yet existing methods remain predominantly research prototypes ill-suited for production-level deployment. We identify a critical "Industry-Academia Gap" hindering real-world application: unpredictable resource consumption from heuristic Gaussian growth, the "sparsity shield" of current benchmarks that rewards hallucination over physical fidelity, and severe multi-sensor data pollution. To bridge this gap, we propose \textbf{YOGO (You Only Gaussian Once)}, a system-level framework that reformulates the stochastic growth process into a deterministic, budget-aware equilibrium. YOGO integrates a novel budget controller for hardware-constrained resource allocation and an availability-registration protocol for robust multi-sensor fusion. To push the boundaries of reconstruction fidelity, we introduce \textbf{Immersion v1.0}, the first ultra-dense indoor dataset specifically designed to break the "sparsity shield." By providing saturated viewpoint coverage, Immersion v1.0 forces algorithms to focus on extreme physical fidelity rather than viewpoint interpolation, and enables the community to focus on the upper limits of high-fidelity reconstruction.  Extensive experiments demonstrate that YOGO achieves state-of-the-art visual quality while maintaining a strictly deterministic profile, establishing a new standard for production-grade 3DGS. To facilitate reproducibility, part scenes of Immersion v1.0 dataset and source code of YOGO has been publicly released. The project link is https://jjrcn.github.io/yogo-project-home/.

\end{abstract}

\section{Introduction}
\label{sec:intro}

The emergence of 3D Gaussian Splatting (3DGS) \cite{kerbl3Dgaussians} has revolutionized radiance field reconstruction with its explicit geometry and real-time rasterization. However, it remains predominantly an academic construct. While excelling on sparse-view benchmarks \cite{barron2022mipnerf360,Knapitsch2017,ling2024dl3dv}, 3DGS struggles with the extreme textures, complex occlusions, and strict resource constraints inherent to production environments like digital twins \cite{jia2023competition,jia2024ropebev,ju2021danet,kang2025sat2realcity,shi2023open,xu2025cruise,xing2025doremi} and autonomous driving \cite{jinrang2023monouni,kong2023dusa,li2024monolss,xia2024vit,chen2023transiff}.

We attribute this limited deployability to a critical Industry-Academia Gap. First, standard 3DGS relies on heuristic densification, rendering the final primitive count an unpredictable byproduct. For edge deployment, this non-determinism necessitates costly trial-and-error parameter tuning. Second, existing sparse-view benchmarks force algorithms to hallucinate missing structures rather than maximize physical fidelity, masking their true modeling capacity. Finally, real-world acquisition inherently suffers from "data pollution"—inconsistencies across heterogeneous sensors with varying exposures and noise profiles.

To bridge these gaps, we introduce \textbf{YOGO (You Only Gaussian Once)}, a system-level framework engineered for deterministic, robust, and production-ready 3D reconstruction. YOGO redefines the 3DGS pipeline through four core contributions:
\begin{itemize}
\item \textbf{Deterministic Budget Control:} We replace heuristic growth with a feedback-loop controller, enforcing strict adherence to hardware-defined primitive budgets. This control can be localized via spatial polygons for extreme fidelity in specific regions of interest.
\item \textbf{Availability-Registration Fusion:} A dynamic protocol that quantifies data "availability," explicitly filtering radiometric inconsistencies and sensor noise to ensure robust fusion from heterogeneous multi-modal sources.
\item \textbf{Solid Optimization Suite:} A collection of targeted enhancements—including Area-Normalized Gradient Accumulation, Maximum Effective Opacity Pruning, and Principal Axis Densification—to preserve high-frequency details and compress model scale.
\item \textbf{Immersion v1.0 Benchmark:} A novel, ultra-dense indoor dataset featuring saturated multi-sensor acquisition. It shifts the evaluation paradigm from sparse-view synthesis to absolute physical fidelity and cross-sensor robustness.
\end{itemize}

Ultimately, YOGO transcends the laboratory prototype, establishing a new standard for resource-controllable, high-fidelity rendering. Extensive evaluations on Immersion v1.0 demonstrate that YOGO achieves state-of-the-art visual quality while maintaining a strictly deterministic profile, proving its efficacy for real-world deployment.

\section{Related Work}
\label{sec:related_work}

\subsection{Evolution of Neural 3D Reconstruction}
Neural Radiance Fields (NeRF) \cite{mildenhall2020nerfrepresentingscenesneural} and Instant-NGP \cite{mueller2022instant} pioneered implicit volumetric rendering. Recently, 3D Gaussian Splatting (3DGS) \cite{kerbl3Dgaussians} shifted the paradigm to explicit point-based rendering, achieving real-time speeds and catalyzing rapid architectural advancements \cite{jia2026panoworld,Li_2024_CVPR,li2026pano2world,liu2025attentiongsinitializationfree3dgaussian,Lyu_2025_ICCV,Yuan_2025_ICCV}. Subsequent works further enhanced geometric modeling and anti-aliasing (SuGaR \cite{guedon2023sugar}, 2DGS \cite{Huang2DGS2024}, Mip-Splatting \cite{Yu2024MipSplatting}), as well as densification efficiency and gradient accumulation (Scaffold-GS \cite{scaffoldgs}, absGS \cite{ye2024absgs}, pixelGS \cite{zhang2024pixelgs}). Despite these improvements, existing methods rely on heuristic densification thresholds, treating reconstruction as an uncontrollable process. This unpredictability in memory footprint hinders strict hardware deployment—a critical gap YOGO bridges via deterministic resource control.

\subsection{Resource-Constrained Optimization}
To mitigate the massive memory footprint of 3DGS, existing literature primarily employs pruning \cite{guedon2023sugar,scaffoldgs,xing2026adaptsplat} or post-training compression techniques \cite{fang2024minisplattingrepresentingscenesconstrained,niedermayr2023compressed,wang2025rt}. However, these "train-then-fix" approaches lack determinism: users cannot pre-define exact hardware budgets (e.g., "exactly 2M Gaussians") prior to training. YOGO diverges by reformulating densification as a controllable equilibrium, enabling one-stage training that strictly adheres to pre-set resource constraints from the outset.

\subsection{3D Reconstruction Datasets}
Traditional benchmarks \cite{barron2022mipnerf360,Knapitsch2017} and larger-scale datasets \cite{dai2017scannet,ling2024dl3dv,wang2026artifactworld} typically feature sparse viewpoint sampling and limited sensor diversity. This under-constrained nature forces algorithms to solve for structure completion, obscuring their intrinsic modeling power. Our Immersion v1.0 dataset addresses this by providing ultra-dense viewpoint saturation and multi-sensor registration, enabling algorithms to focus exclusively on the pursuit of extreme reconstruction quality and physical fidelity.

\begin{figure*}[t]
  \centering
  \includegraphics[width=\linewidth]{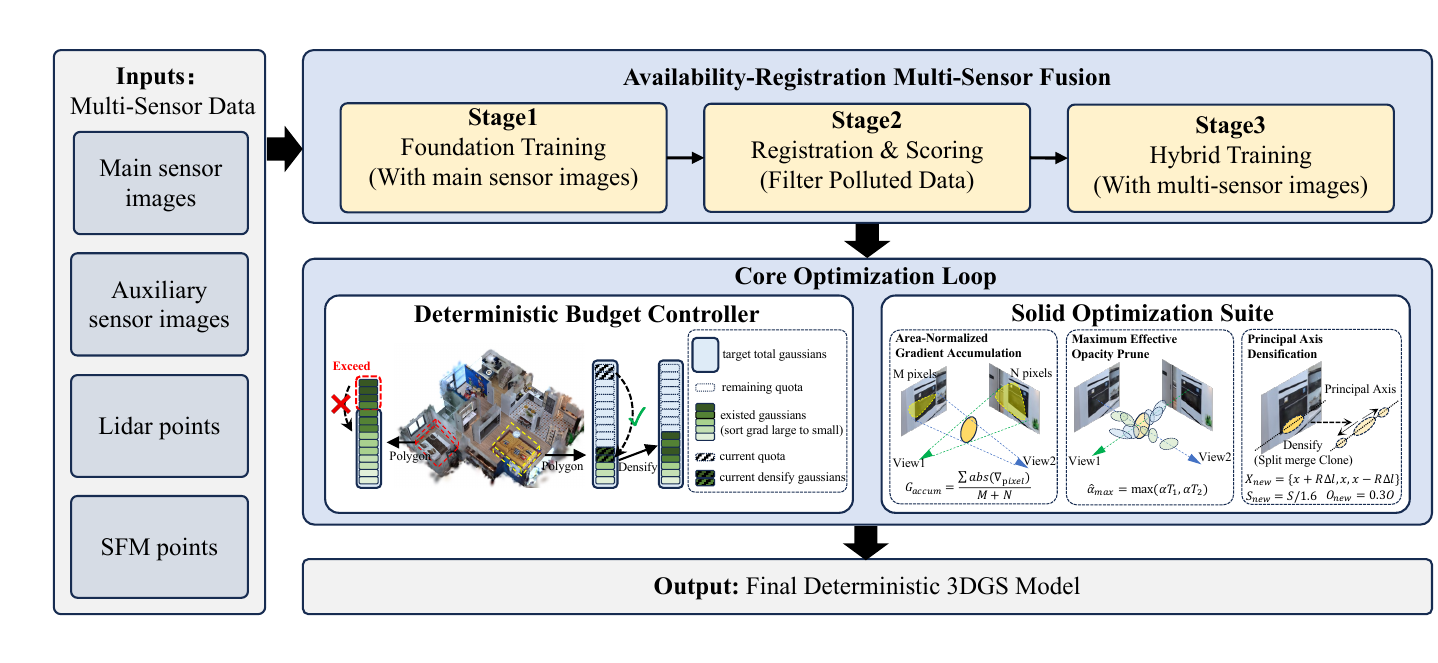} 
  \caption{\textbf{Overview of the YOGO Framework.} The pipeline begins with multi-sensor data undergoing Availability-Registration Multi-Sensor Fusion (Sec.~\ref{Availability_Registration_Multi_Sensor_Fusion}) to filter polluted inputs. Under the deterministic budget controller (Sec.~\ref{Deterministic_Budget_Controller}), the number of Gaussian points at each stage is strictly controlled, which regulates growth based on preset constraints and Polygon regions. The process is enhanced by the Solid Optimization Suite (Sec.~\ref{Solid_Optimization_Suite}) for robust handling of complex textures.}
  \label{fig:architecture}
\end{figure*}

\section{Methodology}
\label{sec:method}
\subsection{Deterministic Budget Controller}
\label{Deterministic_Budget_Controller}
Standard 3DGS pipelines rely on heuristic densification thresholds, leading to unpredictable primitive counts. To facilitate production-level deployment, we introduce the Deterministic Budget Controller (DBC), reformulating stochastic Gaussian growth into a budget-driven process for strict hardware compliance and Region of Interest (ROI) refinement.
\subsubsection{Spatial Partitioning and Formulation}Instead of global unconstrained densification, we partition the scene into $M$ disjoint spatial polygons $\mathcal{P}=\{P_1,\dots,P_M\}$. Each polygon is assigned a target Gaussian budget $N^{target}_{m}$ based on user or hardware constraints. Given a densification schedule with start iteration $S$, end iteration $E$, and interval $D$, the total number of densification events is:$$K=\left\lfloor\frac{E-S}{D}\right\rfloor+1$$
\subsubsection{Budget-Aware Regulation}For each densification event $k \in [1, K]$, we apply a "prune-then-densify" sequence. After pruning, let $P_m$ contain $N^{cur}_{m}(k)$ Gaussians. The remaining budget gap is $\Delta N_m(k)=N^{target}_{m}-N^{cur}_{m}(k)$. To ensure stable convergence, the densification quota for the current step is allocated dynamically:$$N^{densify}_{m}(k)=\max\left(0,\frac{\Delta N_m(k)}{K-k+1}\right)$$
\subsubsection{Importance-Driven Selection}Unlike methods using fixed gradient thresholds, YOGO selects the top $Q=N^{densify}_{m}(k)$ Gaussians with the highest accumulated gradient magnitudes $\|\nabla_{\mu}\|$ within each $P_m$. This mechanism provides two critical advantages:
\begin{itemize}
\item \textbf{Deterministic Predictability:} The final primitive count strictly converges to $\sum N^{target}_{m}$, eliminating the need for iterative parameter tuning.
\item \textbf{Multi-Granular Control:} Adjusting $N^{target}_{m}$ enables highly localized resource allocation, preserving extreme fidelity in complex ROIs while maintaining high compression ratios in backgrounds.
\end{itemize}

\subsection{Availability-Registration Multi-Sensor Fusion}
\label{Availability_Registration_Multi_Sensor_Fusion}
To mitigate data pollution from multi-modal sensor inconsistencies, we propose an Availability-Registration Protocol. Unlike naive hybrid training, this hierarchical approach identifies and filters non-compliant data via a rigorous three-stage pipeline.
\subsubsection{Hierarchical Training Pipeline}Our protocol integrates multi-sensor data through the following sequence:
\begin{itemize}
\item \textbf{Foundation Training:} We establish a baseline 3DGS anchor utilizing data from the primary sensor $\mathcal{D}_{pri}$.
\item \textbf{Radiometric Registration:} We freeze the anchor model and introduce downsampled auxiliary data $\mathcal{D}_{aux}$. To decouple exposure and sensor bias, we optimize a per-view affine transformation for each auxiliary image $I_j$: $C_{trans}=\mathbf{G}_j C_{rend}+\mathbf{b}_j$, where $\mathbf{G}_j \in \mathbb{R}^{3 \times 3}$ and $\mathbf{b}_j \in \mathbb{R}^3$ form the radiometric transformation matrix $\mathbf{A}_j=[\mathbf{G}_j|\mathbf{b}_j]$.
\item \textbf{Hybrid Training:} We filter anomalous auxiliary frames based on a quantitative availability score, jointly optimizing the remaining high-quality data with $\mathcal{D}_{pri}$.
\end{itemize}
\subsubsection{Availability Scoring and Outlier Rejection}An auxiliary sample is deemed "available" if its mapping to the anchor avoids extreme or non-physical transformations. We define the Availability Score $S_j$ as the total deviation of $\mathbf{A}_j$ from the identity matrix $\mathcal{I}=[\mathbf{I}|\mathbf{0}]$:$$S_j=\|\text{diag}(\mathbf{G}_j)-\mathbf{1}\|_{\infty}+\text{avg}(|\text{offdiag}(\mathbf{G}_j)|)+\text{avg}(|\mathbf{b}_j|)$$Here, the three terms explicitly penalize peak channel-wise gain deviations, cross-channel crosstalk, and global luminance bias, respectively. Samples exceeding a strict threshold $S_j>\tau$ are rejected. This formulation effectively isolates significant radiometric distortion and geometric misalignment, ensuring the YOGO engine converges to a high-fidelity representation across heterogeneous data.

\subsection{The Solid Optimization Suite}
\label{Solid_Optimization_Suite}To achieve extreme fidelity, YOGO integrates targeted optimizations for high-frequency textures and robust rendering.
\subsubsection{Area-Normalized Gradient Accumulation}Standard position gradients $\nabla_{\mu}$ often blur texture-dense regions. Inspired by absGS \cite{ye2024absgs} and pixelGS \cite{zhang2024pixelgs}, we propose the Area-Normalized Absolute Gradient $\bar{G}_{accum}$ to ensure unbiased densification:$$\bar{G}_{accum} = \frac{\sum_{i \in \mathcal{V}} \sum_{p \in \Omega_i} |\nabla_{p}|}{\sum_{i \in \mathcal{V}} |\Omega_i|}$$where $\mathcal{V}$ denotes visible views, $\Omega_i$ is the pixel footprint in view $i$, and $\nabla_{p}$ is the pixel-wise gradient. Unlike pixelGS, which favors large Gaussians, our normalization by cross-view pixel count prevents footprint bias. By aggregating per-pixel absolute gradients, we avoid directional signal cancellation, preserving sharp edges. This density-invariant metric precisely targets regions with high reconstruction error.
\subsubsection{Pruning via Maximum Effective Opacity}Standard pruning ($\alpha < \epsilon$) fails to remove heavily occluded primitives. We introduce the Effective Opacity $\hat{\alpha} = \alpha \cdot T$, where $T$ is the accumulated transmittance. At each epoch, we compute the Maximum Per-View Effective Opacity:$$\hat{\alpha}_{max} = \max_{i \in \mathcal{V}} \left( \frac{\sum_{p \in \Omega_i} \hat{\alpha}_p}{|\Omega_i|} \right)$$Primitives with $\hat{\alpha}_{max} < \tau_{opacity}$ are pruned. This eliminates visually redundant Gaussians that possess high intrinsic opacity but contribute zero to the rendering, significantly compressing the model without quality degradation.
\subsubsection{Principal Axis Densification}To recover thin structures and anisotropic textures efficiently, we unify "split" and "clone" operations into a Principal Axis Densification strategy. For a Gaussian exceeding the gradient threshold, we generate a triplet along its major scaling axis. Given scale $\mathbf{s}$, rotation $\mathbf{R}$, and principal axis index $q = \arg\max(\mathbf{s})$, we sample a local perturbation $\Delta \mathbf{l} \sim \mathcal{N}(0, 0.3 s_q) \cdot \mathbf{e}_q$. The new world-space positions are:$$\mathbf{x}_{new} = \{ \mathbf{x} + \mathbf{R}\Delta \mathbf{l}, \mathbf{x}, \mathbf{x} - \mathbf{R}\Delta \mathbf{l} \}$$To conserve energy, scaling and opacity are attenuated by factors of 1/1.6 and 0.3, respectively. This directional expansion outperforms isotropic cloning in complex geometries.

\begin{figure*}[t]
  \centering
  \includegraphics[width=\linewidth]{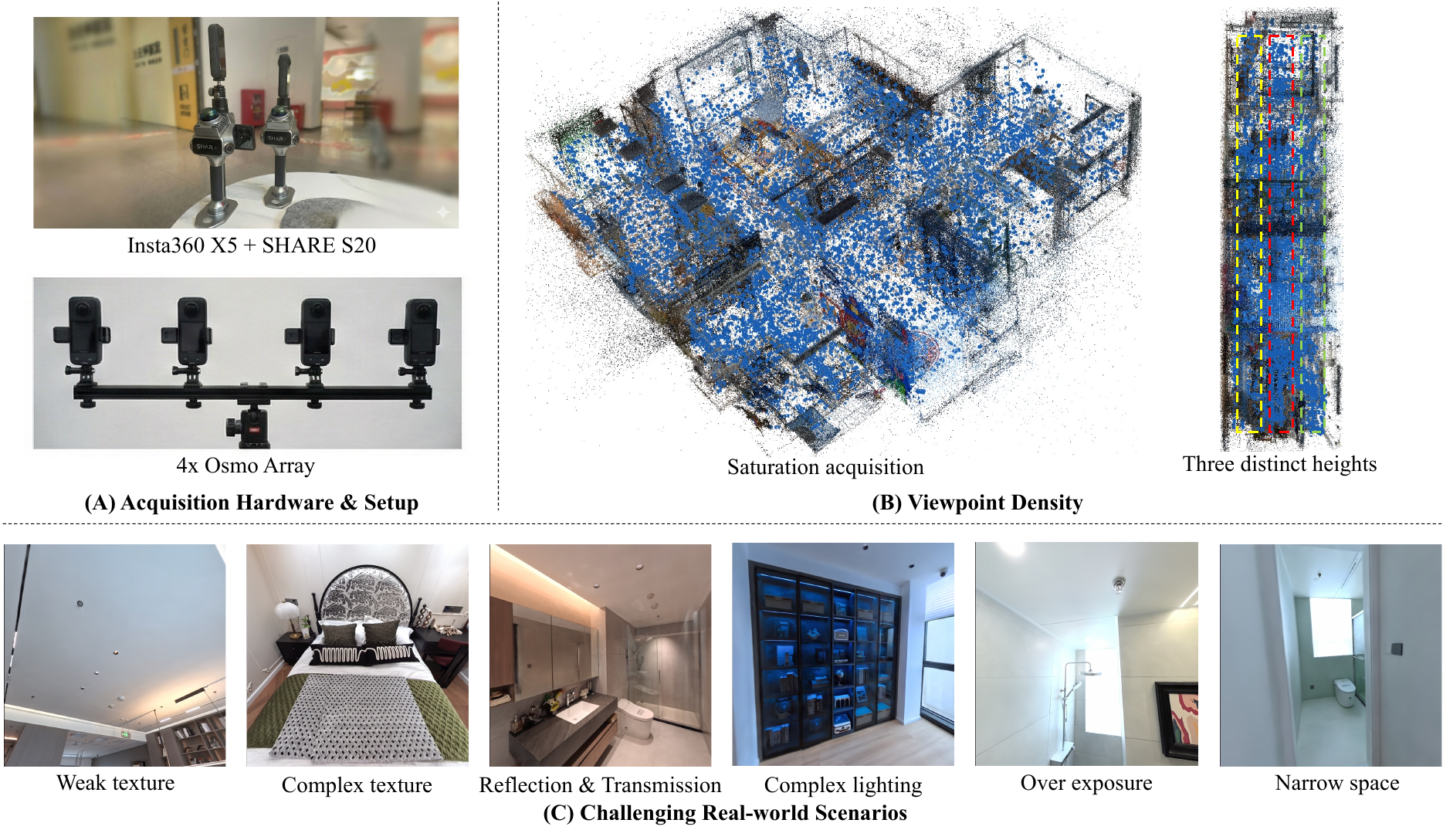}
  \caption{\textbf{Characteristics of the Immersion Dataset.} (A) Heterogeneous multi-sensor capture rigs. (B) Ultra-dense capture that breaks the "sparsity shield." (C) Example frames showing challenging real-world attributes: weak texture, complex texture, high specularity \& transmission, complex lighting, over exposure, and narrow space.}
  \label{fig:dataset}
\end{figure*}

\section{The Immersion v1.0 Dataset}
\label{sec:Immersion}
Existing 3DGS benchmarks \cite{barron2022mipnerf360,Knapitsch2017,yeshwanth2023scannet++} suffer from sparse viewpoint sampling, compelling algorithms to ``hallucinate'' missing structures in under-constrained settings. Immersion v1.0 breaks this ``sparsity shield'' via an ultra-dense, multi-sensor saturated acquisition protocol. By ensuring comprehensive scene coverage, it shifts the evaluation focus from view interpolation to high-fidelity, production-grade physical reconstruction.

\subsection{Heterogeneous Multi-Sensor Acquisition System}
To reflect production-level captures, Immersion v1.0 utilizes a synchronized suite of high-end sensors (Table \ref{table_1}):
\begin{itemize}
\item \textbf{Insta360 X5 Engine}: Captures $7680 \times 3840$ panoramic video (24 fps). Frames are extracted at 2 Hz and re-projected to generate four $960 \times 960$ and two $1280 \times 960$ rectified views per timestamp.
\item \textbf{Osmo High-Dynamic Array}: Four synchronized units capturing $3840 \times 2880$ (25 fps), sampled at 2 Hz and resized to $1440 \times 1080$ for low-noise perspective detail.
\item \textbf{SHARE SLAM S20}: A hybrid LiDAR-Visual system providing dual-lens images ($876 \times 1168$ at 2 Hz) synchronized with a 200k pts/sec LiDAR, yielding geometric ground truth for initial point clouds and SfM calibration.
\end{itemize}

\subsection{Saturated Acquisition Protocol}
We deploy two primary configurations: Insta360+S20 and a 4-Osmo+S20 array. To guarantee complete viewpoint saturation, we execute a \textbf{Six-Loop Trajectory Strategy}: bi-directional (clockwise and counter-clockwise) loops at three distinct heights (High, Medium, Low), supplemented by targeted close-up shots for thin structures. This protocol achieves an unprecedented density, observing every voxel from hundreds of overlapping angles.

\begin{table*}[t]
\centering
\footnotesize
\setlength{\tabcolsep}{1pt}
\begin{tabular}{l|c|c|c|c|c|c|c|c|c}
\hline
\multirow{2}*{Dataset} & \multirow{2}*{Scenes} & Avg. & Avg. & Avg. & Avg. & Avg. & \multirow{2}*{M.S.} & \multirow{2}*{M.R.} & \multirow{2}*{LiDAR}  \\ 
& & Imgs & Init Points & Area & IDSM & Cov. & & & \\
\hline
\hline
Mip-NeRF 360 \cite{barron2022mipnerf360} & 9 & $\sim$200 & $\sim$158K & - & - & - & No & No & No \\
Tanks \& Temples \cite{Knapitsch2017} & 21 & $\sim$300 & $\sim$159K & - & - & - & No & No & No \\
DL3DV \cite{ling2024dl3dv} & 10K & $\sim$340 & $\sim$75K & - & - & - & No & No & No \\
ScanNet$++$ \cite{yeshwanth2023scannet++} & 1K & $\sim$2K & $\sim$161K & $\sim$23 & $\sim$130 & $\sim$16\% & Yes & Rare & Yes \\ 
\textbf{Immersion v1.0} & 7 & \textbf{$\sim$30K} & \textbf{$\sim$2.55M} & \textbf{$\sim$100} & \textbf{$\sim$500} & \textbf{$\sim$72\%} & \textbf{Yes} & \textbf{All} & \textbf{Yes} \\
\hline
\end{tabular}
\caption{Comparison of different datasets. M.S. indicates multi-sensor data; M.R. denotes multi-room environments; LiDAR indicates LiDAR availability.}
\label{table_1}
\end{table*}

\subsection{Dataset Characteristics}
We quantify our dataset's saturation using Image Density per Square Meter (IDSM) and a proposed grid-based \textbf{Scene Coverage} metric. We discretize the scene floor into $0.2m \times 0.2m$ cells. A grid $\mathcal{G}$ is ``effectively occupied'' if it contains $>6$ camera poses. Coverage is the ratio of effectively occupied grids to total grids within the bounding envelope:
$$Coverage = \frac{Count(\mathcal{G}_{occupied})}{Count(\mathcal{G}_{total})}$$

As detailed in Table \ref{table_1}, Immersion v1.0 represents a significant leap in data scale:
\begin{itemize}
\item \textbf{Unprecedented Volume \& Density}: Generating $\sim$30K frames and $\sim$2.55M initial points per scene, our dataset yields a 15--100$\times$ scale increase over benchmarks like ScanNet++ \cite{yeshwanth2023scannet++}, Mip-NeRF 360 \cite{barron2022mipnerf360}, and Tanks \& Temples \cite{Knapitsch2017}. Its 72\% Coverage and 500 IDSM are roughly 4$\times$ higher than ScanNet++, providing a nearly continuous observation of the environment. 
\item \textbf{Fitness for Overfit Training}: Although comprising 7 scenes, this scale is fully sufficient for per-scene overfit training methods like Perceptual-GS \cite{zhou2025perceptualgssceneadaptiveperceptualdensification} and Scaffold-GS \cite{scaffoldgs}, which prioritize data depth over scene breadth.
\item \textbf{Structural Complexity}: Unlike datasets focused on single rooms \cite{yeshwanth2023scannet++,ling2024dl3dv}, Immersion v1.0 encapsulates complex, multi-room (M.R.) environments with complete LiDAR support and multi-sensor (M.S.) synchronization.
\end{itemize}

\begin{table*}[t]
\centering
\begin{tabular}{c|l|c|cccc|c}
\hline
\multirow{2}*{Track} & \multirow{2}*{Method}& \multirow{2}*{Point} & \multicolumn{4}{c|}{Evalutation} & \multicolumn{1}{c}{Test} \\ \cline{4-8}
 & & & PSNR & SSIM & LPIPS & Qalign & Qalign \\
\hline\hline
\multirow{8}*{SSS} & 3DGS \cite{kerbl3Dgaussians}  & 1.46M & 22.49 & 0.8407 & 0.3375 & 2.7290 & 2.6571 \\
 & AbsGS \cite{ye2024absgs} & 3.13M & 21.85 & 0.8214 & 0.3572 & 2.7215 & 2.6076 \\
 & Mip-Splatting \cite{Yu2024MipSplatting} & \colorbox{paperGreen}{1.04M} & 22.87 & 0.8429 & 0.3298 & 2.6522 & 2.5662 \\
 & Scaffold-GS \cite{scaffoldgs} & \colorbox{paperRed}{1.04M} & 24.38 & 0.8527 & 0.3158 & 2.5800 & 2.5197 \\
 & Perceptual-GS \cite{zhou2025perceptualgssceneadaptiveperceptualdensification} & 3.46M & 22.59 & 0.8412 & 0.3234 & 2.8443 & 2.8226 \\
 & YOGO1 & \colorbox{paperYellow}{1.76M} & \colorbox{paperYellow}{25.83} & \colorbox{paperYellow}{0.8674} & \colorbox{paperYellow}{0.3001} & \colorbox{paperYellow}{3.1789} & \colorbox{paperYellow}{3.1764} \\
 & YOGO2 & 3.98M & \colorbox{paperGreen}{25.90} & \colorbox{paperGreen}{0.8695} & \colorbox{paperGreen}{0.2930} & \colorbox{paperGreen}{3.2801} & \colorbox{paperGreen}{3.3053} \\
 & YOGO3 & 5.83M & \colorbox{paperRed}{25.92} & \colorbox{paperRed}{0.8701} & \colorbox{paperRed}{0.2904} & \colorbox{paperRed}{3.3220} & \colorbox{paperRed}{3.3485} \\
\hline
\multirow{5}*{SSD} & 3DGS \cite{kerbl3Dgaussians}  & \colorbox{paperGreen}{2.68M} & 27.50 & 0.8812 & 0.2737 & 3.6292 & 3.6203 \\
 & AbsGS \cite{ye2024absgs} & 4.28M & 27.62 & 0.8823 & 0.2700 & 3.6376 & 3.6404 \\
 & YOGO1 & \colorbox{paperRed}{1.49M} & \colorbox{paperYellow}{27.73} & \colorbox{paperYellow}{0.8870} & \colorbox{paperYellow}{0.2681} & \colorbox{paperYellow}{3.6839} & \colorbox{paperYellow}{3.7142} \\
 & YOGO2 & \colorbox{paperYellow}{3.50M} & \colorbox{paperGreen}{27.81} & \colorbox{paperGreen}{0.8885} & \colorbox{paperGreen}{0.2645} & \colorbox{paperGreen}{3.7380} & \colorbox{paperGreen}{3.7774} \\
 & YOGO3 & 5.64M & \colorbox{paperRed}{27.84} & \colorbox{paperRed}{0.8891} & \colorbox{paperRed}{0.2632} & \colorbox{paperRed}{3.7543} & \colorbox{paperRed}{3.7959} \\
\hline
\multirow{5}*{MSD} & 3DGS \cite{kerbl3Dgaussians}  & \colorbox{paperGreen}{3.14M} & 27.24 & 0.8806 & 0.2719 & 3.6573 & 3.6763 \\
 & AbsGS \cite{ye2024absgs} & 5.33M & 27.34 & 0.8820 & \colorbox{paperYellow}{0.2669} & 3.6565 & 3.6805 \\
 & YOGO1 & \colorbox{paperRed}{1.45M} & \colorbox{paperYellow}{27.57} & \colorbox{paperYellow}{0.8862} & 0.2681 & \colorbox{paperYellow}{3.6816} & \colorbox{paperYellow}{3.7771} \\
 & YOGO2 & \colorbox{paperYellow}{3.45M} & \colorbox{paperGreen}{27.63} & \colorbox{paperGreen}{0.8878} & \colorbox{paperGreen}{0.2643} & \colorbox{paperGreen}{3.7294} & \colorbox{paperGreen}{3.8211} \\
 & YOGO3 & 5.23M & \colorbox{paperRed}{27.69} & \colorbox{paperRed}{0.8883} & \colorbox{paperRed}{0.2629} & \colorbox{paperRed}{3.7571} & \colorbox{paperRed}{3.8426} \\
\hline
\end{tabular}
\caption{Quantitative results on reconstruction quality, comparing our method with state-of-the-art methods, in terms of PSNR↑, SSIM↑ and LPIPS↓. \colorbox{paperRed}{The best}, \colorbox{paperGreen}{second-best}, and \colorbox{paperYellow}{third-best} results are highlighted.}
\label{table_exp}
\end{table*}

\begin{figure*}[t]
  \centering
  \includegraphics[width=\linewidth]{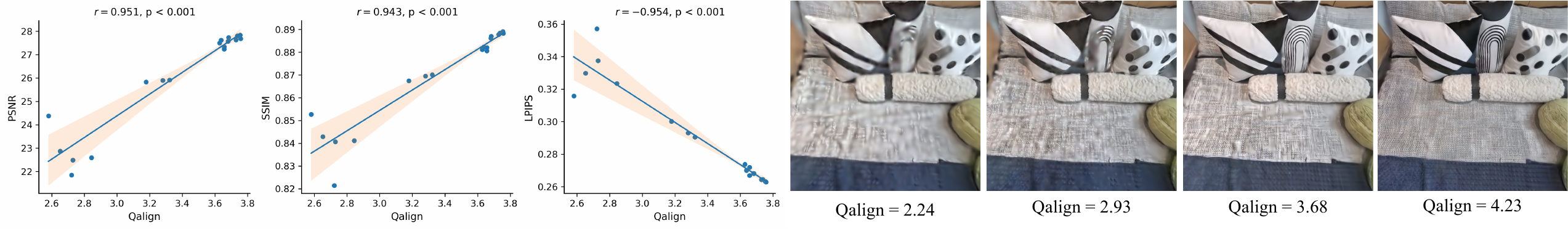}
  \caption{Qalign correlation with PSNR, SSIM, and LPIPS (95\% confidence intervals, p<0.001) alongside example renderings sorted by increasing Qalign. Higher Qalign values consistently correspond to improved perceptual quality, demonstrating its reliability as a no-reference 3DGS metric.}
  \label{fig:qalign}
\end{figure*}

\section{Experiments}
\subsection{Experimental Setup}
\subsubsection{Evaluation Metrics:} We employ standard reconstruction metrics including PSNR, SSIM, and LPIPS \cite{zhang2018perceptual}. However, in ultra-dense real-world captures, pixel-wise metrics often saturate or penalize structurally accurate but slightly misaligned high-frequency details. To robustly assess human-centric visual quality, we introduce Qalign \cite{wu2023qalign}, a perceptual-level metric driven by large multi-modality models.

\subsubsection{Evaluation Data:} We partition the evaluation into two distinct protocols to comprehensively assess reconstruction and generalization:
\begin{itemize}
\item \textbf{Validation Set:} We withhold 100 frames sampled directly from the original capture trajectory. All metrics (PSNR, SSIM, LPIPS, Qalign) are evaluated here to provide standard reference-based benchmarks.
\item \textbf{Roaming Test Set:} To mitigate trajectory-overfitting bias inherent in standard evaluations, we manually navigate the scenes to acquire 200 "roaming" frames from out-of-distribution viewpoints. Lacking paired ground truth, this set rigorously tests structural generalization and is evaluated exclusively via Qalign \cite{wu2023qalign}.
\end{itemize}

\subsubsection{Evaluation Tracks:} To dissect the impact of viewpoint density and sensor fusion, we establish three tracks across our Validation and Test sets:
\begin{itemize}
\item \textbf{Single-Sensor Sparse (SSS):} Downsampled primary sensor data ($\sim$2K images/scene), mirroring the sparsity of conventional datasets \cite{Knapitsch2017,yeshwanth2023scannet++}.
\item \textbf{Single-Sensor Dense (SSD):} Utilizing all primary sensor frames ($>$20K images/scene), serving as the saturated single-modality baseline.
\item \textbf{Multi-Sensor Dense (MSD):} Integrating the full heterogeneous suite ($\sim$30K images/scene), representing the ultimate production-grade challenge.
\end{itemize}

\subsection{Reliability of the Qalign Metric}
\label{Reliability_of_Qalign}

As evidenced in Fig \ref{fig:qalign}, Qalign \cite{wu2023qalign} demonstrates extremely strong correlation with PSNR ($r=0.95, p < 0.001$) and SSIM ($r=0.94, p < 0.001$), while exhibiting strong negative correlation with LPIPS ($r=-0.95, p < 0.001$), indicating high consistency with established perceptual metrics. validating its use as the primary evaluator for the Roaming Test Set where ground truth is inaccessible.

\subsection{Deterministic Budget Controller Analysis}
Unlike heuristic methods, YOGO facilitates explicit control over resource consumption. As shown in Table \ref{table_exp}, we benchmark three budget variants (YOGO1, YOGO2, YOGO3) to analyze the performance-memory trade-off. We observe distinct performance saturation: on the SSS track, upgrading from YOGO1 to YOGO2 (+$\sim$2M points) yields a significant 0.13 Qalign improvement, whereas the transition to YOGO3 (+$\sim$2M points) yields a marginal 0.04 gain. This determinism empowers users to pinpoint the optimal balance between fidelity and computational cost, strictly avoiding redundant memory allocation.

\subsection{State-of-the-Art Comparison on Immersion v1.0}
We benchmark YOGO against leading explicit frameworks: 3DGS \cite{kerbl3Dgaussians}, AbsGS \cite{ye2024absgs}, Mip-Splatting \cite{Yu2024MipSplatting}, Scaffold-GS \cite{scaffoldgs}, and Perceptual-GS \cite{zhou2025perceptualgssceneadaptiveperceptualdensification}. Scaling these baselines to our ultra-dense Immersion dataset poses severe challenges. Methods like Mip-Splatting and Perceptual-GS require global statistical analyses, while Scaffold-GS necessitates exhaustive hyperparameter recalibration. Consequently, due to extreme computational overhead and Out-Of-Memory (OOM) failures, these specific baselines could not be evaluated on the SSD and MSD tracks.

As reported in Table \ref{table_exp}, YOGO consistently outperforms existing architectures. Remarkably, on both SSD and MSD tracks, our most constrained variant (YOGO1, 1.5M points) surpasses the rendering quality of AbsGS (4.28M points), demonstrating superior structural efficiency.

Crucially, evaluating the multi-modal fusion reveals a profound insight into benchmark bias. On the Validation Set, MSD metrics slightly lag behind SSD. However, on the Roaming Test Set, MSD consistently achieves higher Qalign scores. This inversion indicates that while single-sensor models (SSD) readily overfit to the captured trajectory, the integration of auxiliary sensor data (MSD) forces the model to learn a more robust, geometrically complete scene, dramatically enhancing out-of-distribution generalization.

\begin{figure*}[t]
  \centering
  \includegraphics[width=\linewidth]{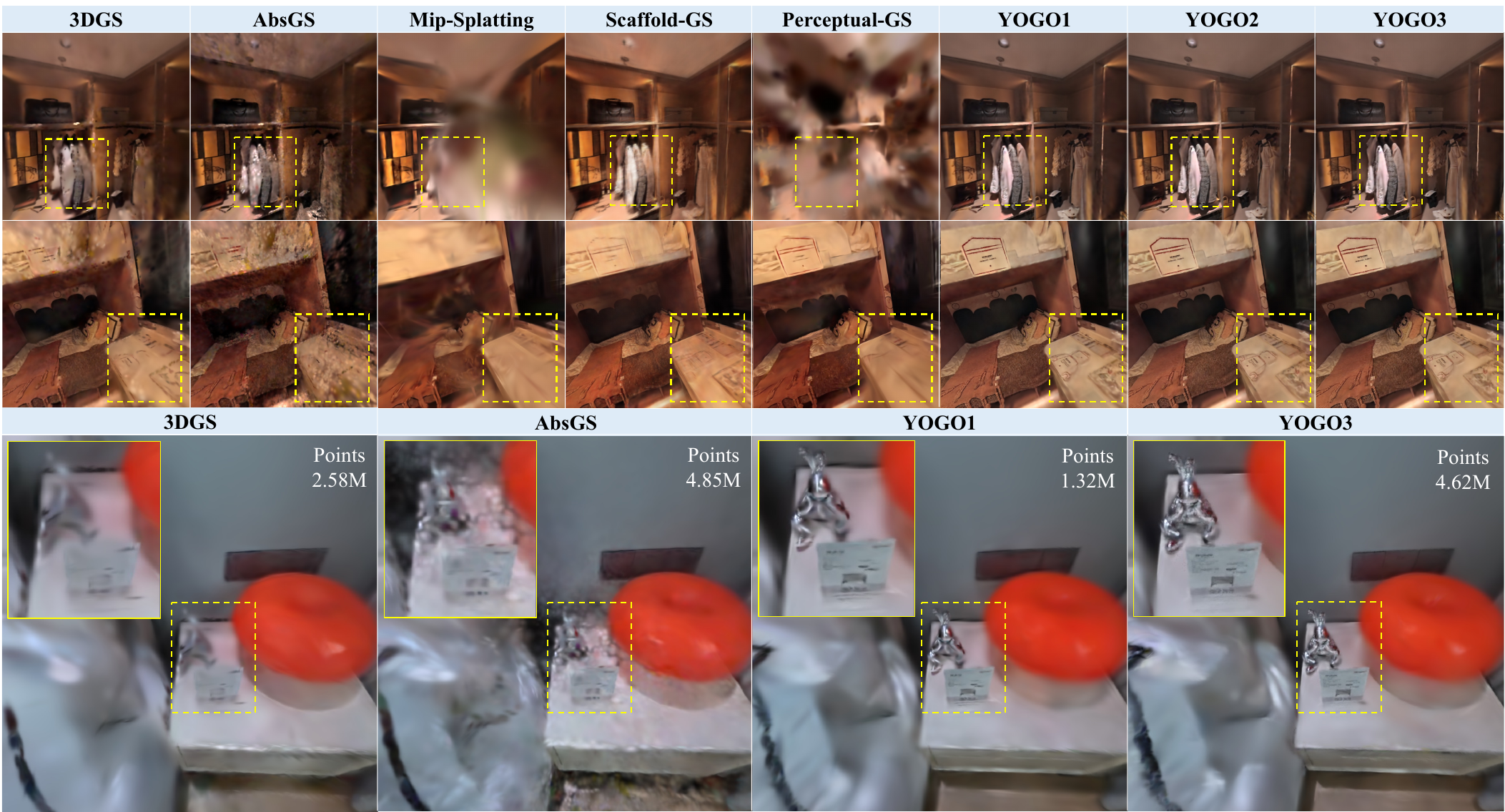}
  \caption{Qualitative comparison across Immersion v1.0 dataset.}
  \label{fig:dataset_Qualitative}
\end{figure*}

\begin{table*}[t]
\centering
\begin{tabular}{c|l|c|c|cccc|c}
\hline
\multirow{2}*{} & \multirow{2}*{Fusion Method} & \multirow{2}*{S20 Images} & \multirow{2}*{Point} & \multicolumn{4}{c|}{Evalutation} & \multicolumn{1}{c}{Test} \\ \cline{5-9}
 & & & & PSNR & SSIM & LPIPS & Qalign & Qalign \\
\hline\hline
(a) & Single Sensor & 0 & 1.49M & \colorbox{paperRed}{27.73} & \colorbox{paperRed}{0.8870} & 0.2681 & \colorbox{paperYellow}{3.6839} & 3.7142 \\
(b) & Direct Fusion & 9769 & \colorbox{paperRed}{1.39M} & 27.34 & 0.8855 & \colorbox{paperGreen}{0.2672} & \colorbox{paperGreen}{3.6934} & 3.7215 \\
(c) & Random Sampling & 3884 & 1.45M & 27.45 & 0.8859 & 0.2685 & 3.6717 & 3.7249 \\
(d) & A.R. with $\tau$=0.1 & 1995 & 1.47M & \colorbox{paperGreen}{27.69} & \colorbox{paperGreen}{0.8865} & 0.2683 & 3.6801 & \colorbox{paperYellow}{3.7368} \\
(e) & A.R. with $\tau$=0.15 & 3884 & \colorbox{paperYellow}{1.45M} & \colorbox{paperYellow}{27.57} & \colorbox{paperYellow}{0.8862} & \colorbox{paperYellow}{0.2681} & 3.6816 &  \colorbox{paperRed}{3.7771} \\
(f) & A.R. with $\tau$=0.3 & 7939 & \colorbox{paperGreen}{1.43M} & 27.37 & 0.8857 & \colorbox{paperRed}{0.2671} & \colorbox{paperRed}{3.6961} & \colorbox{paperGreen}{3.7577} \\
\hline
\end{tabular}
\caption{Ablation studies on Multi-Sensor Fusion Strategies. The metrics are evaluated on the MSD track.}
\label{table_fusion}
\end{table*}

\begin{table*}[t]
\centering
\begin{tabular}{c|c|c|cccc|c}
\hline
\multirow{2}*{$\bar{G}$} & \multirow{2}*{$\hat{\alpha}_{max}$} & \multirow{2}*{PAD} & \multicolumn{4}{c|}{Evalutation} & \multicolumn{1}{c}{Test} \\ \cline{4-8}
 & & & PSNR & SSIM & LPIPS & Qalign & Qalign \\
\hline\hline
 &  &  & 27.50 & 0.8812 & 0.2737 & 3.6292 & 3.6203 \\
\checkmark &  &  & 27.62 & 0.8825 & 0.2702 & 3.6581 & 3.6414 \\
\checkmark & \checkmark &  & 27.69 & 0.8854 & 0.2687 & 3.6670 & 3.7006 \\
\checkmark & \checkmark & \checkmark & 27.73 & 0.8870 & 0.2681 & 3.6839 & 3.7142 \\
\hline
\end{tabular}
\caption{Ablation studies on the Solid Optimization Suite. The metrics are evaluated on the SSD track.}
\label{table_suite}
\end{table*}

\subsection{Ablation Studies}
\subsubsection{Multi-Sensor Fusion Strategies:}
We ablate our Availability-Registration Protocol against naive strategies in Table \ref{table_fusion}: (a) Single Sensor, (b) Direct Fusion, (c) Random Sampling, and (d-f) our A.R. Fusion with varying strictness $\tau$. Direct Fusion degrades performance due to radiometric "data pollution." In contrast, our protocol (A.R. with $\tau=0.15$) successfully isolates severe sensor inconsistencies, systematically filtering outliers while leveraging compliant auxiliary data to boost Test Set generalization.

\subsubsection{The Solid Optimization Suite:}
Table \ref{table_suite} decomposes our optimization modules. Starting from the DBC baseline, we progressively enable Area-Normalized Gradient ($\bar{G}$), Max Effective Opacity Pruning ($\hat{\alpha}_{max}$), and Principal Axis Densification (PAD). Each component yields compounding gains. The synergy between $\bar{G}$ and PAD is particularly potent, effectively preventing blurring in texture-dense regions and recovering anisotropic structures inherent to complex indoor environments.

\section{Conclusion}
We bridge the critical "Industry-Academia Gap" in neural rendering with \textbf{YOGO}, a deterministic 3DGS framework, and \textbf{Immersion v1.0}, an ultra-dense benchmark of complex indoor scenes. By saturating viewpoint coverage, Immersion v1.0 dismantles the "sparsity shield," compelling algorithms to prioritize authentic physical fidelity over view interpolation. Concurrently, YOGO fundamentally transforms heuristic densification into a budget-aware regulation task, enabling arbitrary polygon-level resource allocation and robust heterogeneous sensor fusion.

\noindent\textit{Limitations and Future Work:} A current limitation of Immersion v1.0 is its constrained scene count, a direct consequence of the severe computational overhead required to process massive, saturated multi-sensor captures. We are actively expanding this frontier; Immersion v2.0 is currently underway, featuring over 40 complex environments. Beyond data expansion, our future work will focus on integrating semantic-aware budget allocation into the DBC and extending YOGO to unbounded dynamic environments for autonomous navigation.

%
%
%
\bibliographystyle{splncs04}
\bibliography{mybibliography}

\end{document}